\pdfoutput=1
\documentclass[11pt]{article}

\usepackage[final]{acl}

\usepackage{caption}
\usepackage{float} % 允许使用 [H] 参数

\usepackage{placeins}

\usepackage{times}
\usepackage{latexsym}

\usepackage[T1]{fontenc}
\usepackage[utf8]{inputenc}

\usepackage{microtype}

\usepackage{inconsolata}

\usepackage{graphicx}
\usepackage{booktabs}
\usepackage{multirow}

\usepackage{xcolor}
\usepackage{ifthen}
\newboolean{isdraft}
\setboolean{isdraft}{false}  % Set to false for final version

\ifthenelse{\boolean{isdraft}}{
    \newcommand{\changed}[1]{\textcolor{purple}{#1}}
}{
    \newcommand{\changed}[1]{#1}  % No marking in final version
}
\title{Feel the Difference? A Comparative Analysis of Emotional Arcs in Real and LLM-Generated CBT Sessions}

\author{
  Xiaoyi Wang\textsuperscript{1}, 
  Jiwei Zhang\textsuperscript{1}, 
  Guangtao Zhang\textsuperscript{2*}, 
  Honglei Guo\textsuperscript{3*} \\
  \textsuperscript{1}Department of Computer Science, Shantou University, Shantou, Guangdong, China\\
  \textsuperscript{2}Department of Automation, Tsinghua University, Beijing, China \\
  \textsuperscript{3}BNRIST, Tsinghua University, Beijing, China \\
}

\begin{document}
\maketitle
\renewcommand{\thefootnote}{\fnsymbol{footnote}} % Use symbols for footnotes
\footnotetext[1]{Corresponding author}
\renewcommand{\thefootnote}{\arabic{footnote}}   % Reset back to numbers if needed
\begin{abstract}
Synthetic therapy dialogues generated by large language models (LLMs) are increasingly used in mental health NLP to simulate counseling scenarios, train models, and supplement limited real-world data. However, it remains unclear whether these synthetic conversations capture the nuanced emotional dynamics of real therapy. In this work, we introduce RealCBT, a dataset of authentic cognitive behavioral therapy (CBT) dialogues, and conduct the first comparative analysis of emotional arcs between real and LLM-generated CBT sessions. We adapt the Utterance Emotion Dynamics framework to analyze fine-grained affective trajectories across valence, arousal, and dominance dimensions. Our analysis spans both full dialogues and individual speaker roles (counselor and client), using real sessions from the RealCBT dataset and synthetic dialogues from the CACTUS dataset. We find that while synthetic dialogues are fluent and structurally coherent, they diverge from real conversations in key emotional properties: real sessions exhibit greater emotional variability, more emotion-laden language, and more authentic patterns of reactivity and regulation. Moreover, emotional arc similarity remains low across all pairings, with especially weak alignment between real and synthetic speakers. These findings underscore the limitations of current LLM-generated therapy data and highlight the importance of emotional fidelity in mental health applications. To support future research, our dataset RealCBT is released at \url{https://gitlab.com/xiaoyi.wang/realcbt-dataset}.
\end{abstract}

\section{Introduction}
Mental disorders pose a major global health challenge, affecting nearly one in eight individuals worldwide~\cite{WHOMentalDisorders}. As demand for mental health services continues to rise, a significant barrier remains: the shortage of trained counselors. To help bridge this gap, large language models (LLMs) such as ChatGPT and Gemini are increasingly explored as conversational agents capable of simulating therapeutic interactions and supporting clients in need. The effectiveness of these systems, however, hinges heavily on the quality and diversity of training data---particularly, counseling dialogues grounded in real-world counseling practice.

Unfortunately, access to real counseling dialogues remains extremely limited due to stringent privacy, ethical, and legal constraints. This scarcity of authentic data has led to the widespread adoption of synthetic therapy dialogues generated by LLMs. These synthetic interactions are now commonly used in mental health NLP applications for training, evaluation, and scenario simulation. For example, datasets like CACTUS~\cite{lee-etal-2024-cactus} incorporate psychological principles from Cognitive Behavioral Therapy (CBT) to structure model-generated conversations. While these synthetic dialogues are often fluent and structured around known therapeutic techniques, it remains unclear whether they capture the nuanced emotional dynamics that characterize real counseling sessions.

Emotion plays a central role in both diagnosis and treatment. In CBT, client emotions reveal cognitive distortions, inform intervention strategies, and signal therapeutic progress. A meaningful session often involves shifts in emotional state, such as the alleviation of distress or the emergence of insight. These emotional trajectories, or \textit{emotional arcs}, offer a powerful lens through which to assess the depth, authenticity, and therapeutic quality of a dialogue. It is thus important to understand the extent to which synthetic therapy conversations mirror the emotional dynamics observed in real sessions.

Prior work has demonstrated that emotional arc analysis can yield rich insights into narrative structure across various domains, including novels~\cite{vishnubhotla2024emotion, ohman-etal-2024-emotionarcs, teodorescu-mohammad-2023-evaluating, MOHAMMAD2012730}, social media posts~\cite{vishnubhotla-mohammad-2022-tusc}, and movie scripts~\cite{hipsonEmotionDynamicsMovie2021}. However, emotional arcs remain largely unexplored in the context of psychotherapy, and no prior study has directly compared real and synthetic counseling dialogues from an emotional dynamics perspective.

To address this gap, we curate RealCBT, a dataset of authentic cognitive behavioral therapy dialogues transcribed from public video-sharing platforms (e.g., YouTube and Vimeo). 
We then conduct a comparative emotion trajectory analysis between these real-world sessions and synthetic dialogues from the CACTUS dataset~\cite{lee-etal-2024-cactus}. 
We examine emotional arcs from two perspectives: (1) \textit{The Emotion Dynamics of Real vs. Synthetic Therapy Dialogues}: How does the emotion change in real and LLM-generated CBT dialogues: for entire dialogues, for counselors, and for clients? 
(2) \textit{Emotion Arc Alignment Across Speaker Pairs}: To what extent do emotional trajectories align within and across speaker types (specifically among real–real, synthetic–synthetic, and real–synthetic pairings) for both counselors and clients?

To answer these questions, we conduct the first in-depth comparison of emotional dynamics between real and LLM-generated CBT dialogues. 
We use the \changed{well-established} Utterance Emotion Dynamics (UED) framework
~\cite{teodorescu-mohammad-2023-evaluating, hipsonEmotionDynamicsMovie2021}, \changed{adapted specifically for CBT dialogues,} to measure emotion trajectories over time within counselor–client interactions. 
\changed{Our goal is both to compare real and LLM-generated CBT dialogues and to establish an empirical benchmark for assessing whether synthetic sessions capture the nuanced emotional dynamics of real therapy.
This benchmark provides an important foundation for future work exploring more contextualized and neural emotion models.}

Our analysis reveals that although synthetic dialogues are structurally fluent, they diverge from real sessions across several key emotional dimensions. 
Real therapy sessions display greater emotional variability, more emotion-laden language, and more authentic patterns of emotional reactivity and regulation. 
\changed{Emotional arc correlations are low across all pair types, though synthetic–synthetic pairs show slightly higher alignment than real–real or real–synthetic pairs. }

These findings suggest that while current LLM-generated dialogues may capture surface-level therapeutic cues, they fall short in reproducing the nuanced, co-constructed emotional flow that characterizes real-world therapy, especially in the client role.

Our contributions are threefold:
\begin{itemize}
    \item \textbf{RealCBT dataset}: We curate and release RealCBT, one of the first publicly available datasets of authentic CBT dialogues.
    \item \textbf{Empirical benchmark}: We establish an empirical benchmark for comparing emotional dynamics in real vs. synthetic therapy dialogues, using the UED framework adapted for CBT.
    \item \textbf{Comparative insights}: We identify key divergences between LLM-generated and real CBT sessions, highlighting where current models fail to capture the variability, regulation, and alignment of authentic therapeutic emotion trajectories.
\end{itemize}

%1. Counseling application in different forms have been widely used. Behind of that, are models
%2. The performance of such applications are highly dependent on datasets.
%3. However, the dataset is very limited due to privacy and ethical issues.
%4. Synthetic therapy dialogues generated by LLMs
%5. CACTUS is such an example. Although …., it remains unclear that 
%6. In CBT, emotion is essential for counselors to provide interventions. Unclear whether 
%7. Prior work in novels and movie dialogues, such emotion detection method is valid. But no work in compare real vs synthetic dialogues
%8. In this paper, we analyze …
%9. Our Goal is to ….
%10. Results
%11. Contributions: identify discrepancies, evaluate alignments, offer guidance for improving LLM generated dialogues
%\input{../_sections/02_related_work}
%\input{../_sections/03_experiments}
\section{Related Work}
\subsection{Cognitive Behavior Therapy}
Cognitive Behavioral Therapy (CBT) is a widely adopted, evidence-based form of psychotherapy used to treat a variety of psychological disorders, including depression, anxiety, and addiction~\cite{beck2011cognitive}. 
The core of CBT aims to help clients identify and challenge maladaptive thought patterns, and subsequently restructure these cognitive distortions to promote more realistic and adaptive thinking, ultimately facilitating emotional and behavioral change~\cite{carroll2017cognitive, longmore2007we}. 
Given its structured format and demonstrated effectiveness, CBT has increasingly been adapted into virtual agents to support individuals with limited access to in-person therapy. %need citations for LLM-based virtual agents
\subsection{Synthetic Mental Health Counseling Dialogue Datasets}
Training CBT-based conversational models requires high-quality therapy dialogue datasets. 
However, such data are difficult to obtain due to privacy, ethical, and legal constraints. 
To address this limitation, recent studies have explored the use of LLMs to generate synthetic therapy dialogues that can supplement scarce real-world data. 
Earlier work~\cite{sharma-etal-2023-cognitive, maddela-etal-2023-training, sun-etal-2021-psyqa} has focused on generating single-turn counseling dialogues based on CBT or other psychological strategies. 
More recent research has shifted toward multi-turn generation that better simulates the interactive nature of real therapy sessions~\cite{lee-etal-2024-cactus, xiao-etal-2024-healme, qiu-etal-2024-smile}. 
Among these, CACTUS (CBT-augmented Counseling Chat Corpus) stands out as a publicly available multi-turn dataset designed to capture the flow and depth of CBT-style conversations~\cite{lee-etal-2024-cactus}. 
It uses LLMs to simulate counselor–client interactions guided by CBT theory and therapeutic intent. 
Given its theoretical grounding and multi-turn structure, we adopt CACTUS as the synthetic benchmark dataset for comparison with real-world CBT dialogues in this study. 
Although CACTUS has been evaluated favorably on key metrics such as helpfulness and empathy, recent work~\cite{lee2025realtalk21dayrealworlddataset} raises concerns that LLM-generated therapy dialogues may lack the richness and diversity of emotional expression observed in real interactions. 
This underscores the need to better understand how synthetic dialogues compare to real-world CBT conversations, particularly in terms of emotional dynamics, to guide the design of more realistic and effective counseling models.

\subsection{Lexicon-based Emotion Dynamics Computation}
Emotion dynamics is a psychological framework for measuring how an individual's emotional state evolves over time~\cite{vishnubhotla2024emotion, KUPPENS201722, Hollenstein2015ThisTI}. 
In NLP, lexicon-based approaches are among the most widely used techniques for modeling emotion trajectories in spoken or written language. 
Prominent resources include LIWC~\cite{Tausczik2010Words}, WordNet-Affect~\cite{bobicev2010emotions}, SentiWordNet~\cite{baccianella2010sentiwordnet}, VADER~\cite{hutto2014vader}, and the NRC Emotion Lexicons~\cite{mohammad-2018-obtaining}. 
These lexicons provide emotion-related scores at the word level, enabling the quantification of emotional trends across text.

Lexicon-based methods have been validated in various contexts and shown to achieve high reliability in capturing emotional dynamics~\cite{ohman-etal-2024-emotionarcs, teodorescu-mohammad-2023-evaluating, ohman2021validity}. 
In this study, we adopt a lexicon-based approach, specifically using the NRC Valence, Arousal, and Dominance (VAD) Lexicon~\cite{mohammadNRCVADLexicon2025, mohammad-2018-obtaining} in conjunction with the Utterance Emotion Dynamics (UED) framework~\cite{hipsonEmotionDynamicsMovie2021}. 
This combination allows us to compute the emotional valence, arousal, and dominance at the utterance level, and to model how these emotional states change over the course of each counseling session.

Previous research has successfully applied similar methods to track emotional arcs in literary novels~\cite{vishnubhotla2024emotion, teodorescu-mohammad-2023-evaluating, MOHAMMAD2012730}, social media narratives~\cite{vishnubhotla-mohammad-2022-tusc}, and movie scripts~\cite{hipsonEmotionDynamicsMovie2021}. 
We extend this line of work by applying NRC and UED metrics to therapeutic dialogues, allowing us to compare emotional progression in real and synthetic CBT conversations.
\section{RealCBT: Real-world CBT Dialogue Dataset}
In this section, we describe the process of curating the \textbf{RealCBT} dataset.  
\subsection{Data Collection}
Due to privacy, ethical, and legal constraints, publicly available CBT dialogue datasets are extremely limited. 
To enable comparison with LLM-generated data, we collected real-world CBT Dialogues by identifying video clips from public video-sharing platforms such as YouTube and Vimeo. 
We restricted our search to videos explicitly labeled as CBT-based counseling sessions. 
For a detailed description of the selection criteria, please see Appendix~\ref{sec:realcbt_criteria}.
In total, we collected 76 video-based CBT dialogues that met these criteria. 
Detailed summary statistics of \textbf{RealCBT} is provided in Table~\ref{tab:session_stats_76} in Appendix~\ref{sec:realcbt_summary}.
\subsection{Preprocessing and Transcription}
All videos were first converted into a standard MP4 format and then preprocessed to remove non-conversational, user-generated content such as introductory titles, animations, and narration. 
We used the Transkriptor\footnote{\url{https://transkriptor.com/}} service to generate initial transcripts for each video. 
To ensure transcript quality and temporal alignment, each transcript was manually reviewed and corrected to accurately reflect the spoken dialogue in the video. 
Additionally, we reviewed metadata associated with each video (e.g., the number of views, likes, and information about the channel owner) to verify that the content originated from credible sources. 
These indicators collectively suggest that the selected videos represent professionally conducted CBT sessions. 
\subsection{Metadata Annotation}
%While CACTUS provides limited metadata, such as client problem type, client gender, and a predefined overall client attitude toward therapy, it does not include explicit annotations. 
%To align our real-world dataset with CACTUS, 
Similar to CACTUS, we annotated each dialogue with the following attributes from video metadata: (1) client problem, (2) client gender, and (3) overall client attitude toward the session. 
We employed three state-of-the-art language models (i.e., ChatGPT-4o Mini, Grok-v3, and Gemini 2.0 Flash) to independently infer these tags for each dialogue. 
A majority voting scheme was then used to determine the final label for each attribute. 
To evaluate the accuracy of this automated annotation approach, we manually reviewed a randomly selected subset comprising 30\% of the dialogues. 
The automated predictions were consistent with the human annotations in this subset, yielding an estimated labeling accuracy of 100\%. 
Detailed distribution of the metadata of \textbf{RealCBT} is provided in Table~\ref{tab:distribution_realcbt_76} in Appendix~\ref{sec:realcbt_summary}.
\section{Emotion Dynamics Computation}
\label{sec:emotion_dynamics_computation}
We adapt emotion dynamics methods from prior work by Mohammad and colleagues~\cite{vishnubhotla2024emotion, teodorescu-mohammad-2023-evaluating, hipsonEmotionDynamicsMovie2021}, who demonstrated robust approaches for capturing emotional arcs in narratives and movies. 
Specifically, we adopt the Utterance Emotion Dynamics (UED) framework to compute emotion metrics from sequences of utterances, treating each role (counselor or client) as a separate character trajectory. 

Given that CBT sessions are considerably shorter than novels or long-form narratives, we adjust the emotion arc computation accordingly. Following prior work~\cite{hipsonEmotionDynamicsMovie2021, vishnubhotla2024emotion}, we apply a rolling window of 10 words, advancing one word at a time, to compute fine-grained emotion values across each dialogue. At each windowed step, we estimate the emotional state using version 2 of the NRC Valence, Arousal, and Dominance (VAD) Lexicon~\cite{mohammadNRCVADLexicon2025}, which provides word-level emotion scores for the three affective dimensions: valence (V), arousal (A), and dominance (D). Each V/A/D score ranges from –1 to 1, where 1 indicates a strong positive association and –1 indicates a strong negative association with the corresponding affective dimension.

From the resulting VAD time series, we derive a set of UED metrics~\cite{hipsonEmotionDynamicsMovie2021} that capture the shape and character of emotion progression in therapy dialogues: (1) \textit{Emotion Mean}: The average V/A/D score across a dialogue or speaker's arc; (2) \textit{Emotion Variability}: The standard deviation of V/A/D scores, representing emotional fluctuation or richness; (3) \textit{Average Displacement Length}: Measures the average number of emotion words spoken more or less than usual. Longer displacements suggest stronger or more persistent emotional shifts; (4) \textit{Emotion Rise Rate}: Captures how quickly and intensely a speaker shifts into an emotional state (emotional reactivity or sensitivity); (5)\textit{Emotion Recovery Rate}: Indicates how quickly a speaker returns to their baseline emotional state after a shift (emotion regulation). 

Based on the UED metrics and NRC VAD Lexicon, we compute three types of emotional trajectories per CBT session: (1) the overall dialogue-level arc, (2) the counselor's arc, and (3) the client's arc. 
These arcs are computed separately for both real and synthetic CBT sessions, allowing us to compare matched pairs across datasets (i.e., real vs. synthetic dialogues, real vs. synthetic counselors, and real vs. synthetic clients). 

\section{Experimental Setup}
\begin{figure*}[t]
\centering
  \includegraphics[width=\textwidth]{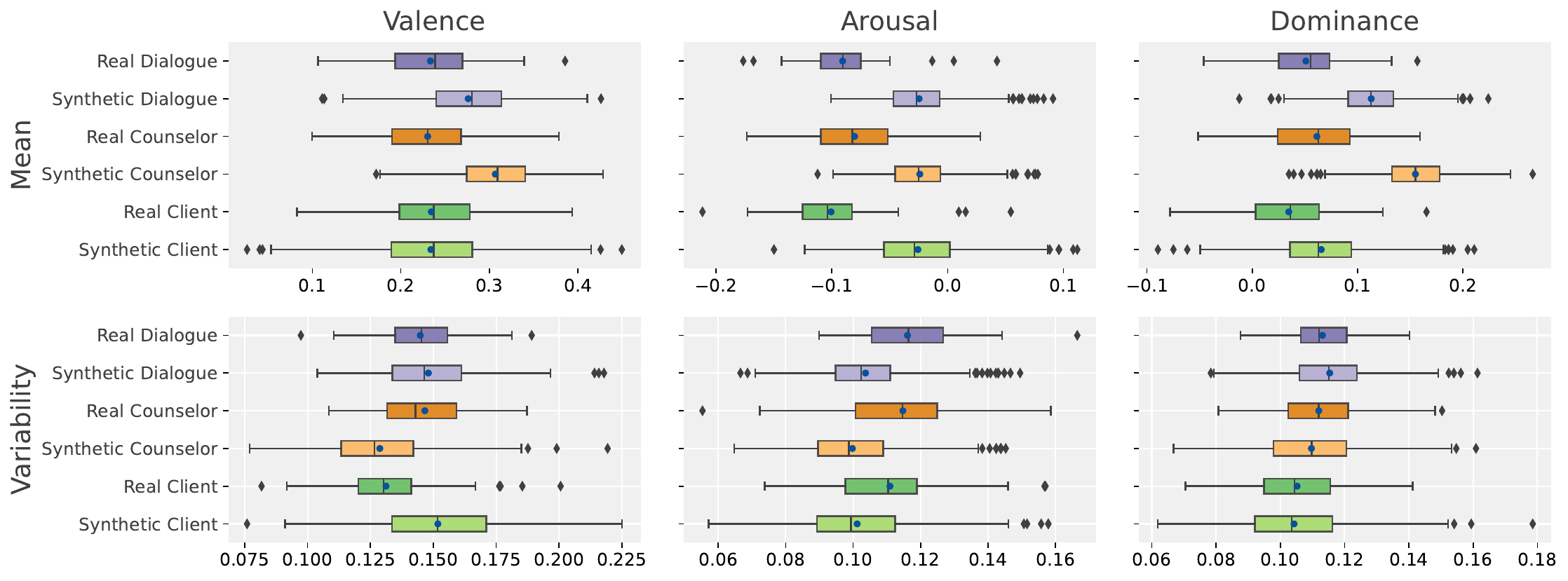}
  \caption{Boxplots showing the distributions of the mean and variability for each of the three affective dimensions across three comparisons: Real vs. Synthetic Dialogues, Real vs. Synthetic Counselors, and Real vs. Synthetic Clients.}
  \label{fig:q1_boxplot}
\end{figure*}

We conduct our study using two datasets. CACTUS consists of 31,564 dialogues with broadly balanced distributions across multiple attributes, making it well suited for large-scale statistical analysis. In contrast, RealCBT comprises 76 dialogues collected from real counseling sessions. Although smaller in scale, RealCBT captures authentic therapeutic interactions and naturally reflects demographic and attitudinal patterns found in practice.

To ensure a fair and meaningful comparison between the two datasets, we adopt a problem-focused sampling strategy. Specifically, we focus on the three most frequent client concerns in RealCBT, including anxiety and fear (25 dialogues), self-esteem and confidence issues (19), and relationship-related concerns (14). Together, these account for 76.3\% of the dataset (58 out of 76 dialogues). For detailed summary statistics and distributional information, see Table~\ref{tab:session_stats_58} and Table~\ref{tab:distribution_realcbt_58} in Appendix~\ref{sec:realcbt_summary}. We then sample synthetic dialogues from CACTUS to match this distribution.

While CACTUS is broadly balanced, RealCBT reflects natural demographic and attitudinal skews (e.g., a higher proportion of female clients and generally positive orientations toward therapy). Such contextual characteristics make strict subgroup balancing less stable, further motivating our problem-focused design.

%CACTUS consists of 31,564 dialogues with roughly balanced distributions across attributes. In contrast, our real-world CBT dataset is relatively small and exhibits imbalance, particularly in terms of client gender and overall attitude toward therapy. Among the 76 collected dialogues, 84.2\% of clients are female (64 out of 76), and 90.8\% express a positive attitude toward the counseling process (69 out of 76). Only 12 dialogues feature male clients, and just 2 involve clients with a negative attitude—making these subgroups too sparse to support reliable statistical comparisons.

%While it is theoretically possible to match gender and attitude proportions when sampling from the synthetic dataset, the severe imbalance in the real dataset would render such comparisons statistically unreliable and potentially misleading.

%To ensure a fair and meaningful comparison, we instead adopt a problem-focused sampling strategy. We focus exclusively on the top three most frequent client concerns in our dataset: anxiety and fear (25 dialogues), self-esteem and confidence issues (19), and relationship-related concerns (14). Together, these categories account for 76.3\% of the dataset (58 out of 76 dialogues). For a detailed description of the summary statistics and distribution of the top three most frequent client concerns in \textbf{RealCBT}, please see Table~\ref{tab:session_stats_58}  and Table~\ref{tab:distribution_realcbt_58} in Appendix A. We sample synthetic dialogues from CACTUS to match this distribution.

To enhance the robustness of our analysis, we repeat the sampling process 10 times without replacement, generating 10 non-overlapping subsets of synthetic dialogues. 
All quantitative results are averaged across these runs to account for sampling variability.

\section{Experimental Results}
\subsection{The Emotion Dynamics of Real vs. Synthetic Therapy Dialogues}
We analyze and compare the following UED metrics across valence, arousal, and dominance: Emotion Mean, Variability, Displacement Length, Rise and Recovery Rates. 
\changed{Detailed definitions and computation methods for each metric are provided in Section~\ref{sec:emotion_dynamics_computation}.} 
To assess statistical significance between real and synthetic groups, we apply the Mann–Whitney U test ($\alpha = 0.05$) to compare real sessions and each of 10 independently sampled synthetic datasets. 
For robustness, we report median p-values and corresponding effect sizes ($r$).  
Figure~\ref{fig:q1_boxplot} presents boxplots illustrating the distributions of Emotion Mean and Variability for each role (i.e., dialogue, counselor, and client) in both real and synthetic sessions. 
Blue dots indicate the aggregated Emotion Mean and Variability for each affective dimension, summarizing overall emotional trends. 
Detailed summary statistics for Emotion Mean, Emotion Variability, and all UED metrics are provided in Table~\ref{tab:vad-stats} - \ref{tab:dominance-stats} in Appendix~\ref{sec:ued_metrics}. 
\subsubsection{Emotion Mean}
As shown in Figure~\ref{fig:q1_boxplot}, synthetic speakers (whole dialogue, counselor, and client) generally sound more emotionally elevated or intense than real human speakers. 
An exception is observed in valence, where real and synthetic clients show similar scores. 
The pairwise comparisons support this observation. 
For entire dialogues (as a whole speaker), we found significant differences in arousal in 90\% of cases (median p-value $<$ 0.001, $r$ = 0.24). 
For counselors, we observed significant differences in valence in all comparisons (median p-value $<$ 0.001, $r$ = 0.36). 
For clients, we found significant differences in arousal in 70.0\% of comparisons (median p-value $<$ 0.05, $r$ = 0.23) and in valence in 90\% (median p-value $<$ 0.05, $r$ = 0.26). 
\subsubsection{Emotion Variability}
For entire CBT dialogues, real sessions exhibited significantly greater variability in arousal than synthetic ones, indicating more fluctuation or richness in real speakers.  
The Mann–Whitney U tests showed significant differences, with a median p-value $<$ 0.0001 and $r$ = 0.84, indicating a strong effect. 
In contrast, valence and dominance variability were similar between real and synthetic dialogues, with only partial significant differences (i.e., 5/10 in dominance and 7/10 in valence).

For counselors, real speakers also showed significantly greater variability in both arousal and valence compared to synthetic counselors. 
For arousal, all comparisons were significant (median p-value $<$ 0.0001, $r$ = 0.41). 
For valence, all comparisons were significant (median p-value $<$ 0.00001, $r$ = 0.42). 
For dominance, real counselors tended to exhibit greater variability, though the differences were not consistently significant across the comparisons. 

For clients, synthetic sessions tended to show greater variability in valence, while real clients showed slightly more variability in arousal. 
However, these differences were only partially significant (i.e., 6/10 for arousal and 1/10 for valence). 
Dominance variability was similar across both groups, with no significant differences observed. 
\subsubsection{Displacement Lengths}
We found that speakers in real dialogues talk significantly more emotion words than the ones in synthetic dialogues in both arousal and dominance. 
In particular, for arousal, all comparisons showed significant differences, with a median p-value $<$ 0.0001 and $r$ = 0.35. 
For dominance, 70\% of comparisons were significant, with a median p-value $<$ 0.05 and $r$ = 0.23. 
In contrast, both real and synthetic speakers uttered similar amount of emotion words in dominance with minimal difference---only 1 out of 10 comparisons reaching significance. 

Real counselors tend to speak more emotion words than synthetic counselors. 
70\% of comparisons showed significant differences in both arousal and valence. 
The median p-values were 0.02 (arousal) and 0.007 (valence), with average effect sizes of 0.23 in both cases. 
In contrast, dominance minimal showed significant differences in only 20\% of comparisons, indicating similar amount of emotion words uttered by both real and synthetic counselors. 

Real clients utter more emotion words than synthetic ones in the dimensions of arousal and dominance. 
Arousal showed consistent differences across all comparisons, with a median p-value $<$ 0.001 and $r$ = 0.335. 
Dominance showed partial significant differences in 70\% of comparisons, with a median p-value $<$ 0.05 and $r$ = 0.24. 
In contrast, valence showed no significant differences across any comparisons. 
\begin{table*}[t]
    \centering
    \small
    \resizebox{\textwidth}{!}{%
    \begin{tabular}{p{1.1cm}p{6.6cm}p{6.6cm}}
    \toprule
    \textbf{Arousal Level} & \textbf{Real Counselor Snippet} & \textbf{Synthetic Counselor Snippet} \\
    \midrule
    \multirow{5}{*}{\textbf{High}} 
        & "It's completely understandable to feel overwhelmed! You've taken a huge step today—this is a big deal, and you're doing really well." 
        & "Wow! That's amazing progress! You should be really proud of yourself—this is wonderful!" \\
        & \textit{Explanation}: Demonstrates emotional reappraisal and support, key to emotion regulation. 
        & \textit{Explanation}: General praise, not a deliberate attempt to regulate the client's emotional state. \\
    \midrule
    \multirow{4}{*}{\textbf{Medium}} 
        & "That's good to hear. Can you walk me through what happened when you started to feel anxious?" 
        & "That's interesting. Let's explore that feeling more." \\
        & \textit{Explanation}: Shows calibrated empathy by gently probing the client's state with a tailored follow-up. 
        & \textit{Explanation}: It's vague and not clearly grounded in what the client just said. \\
    \midrule
    \multirow{4}{*}{\textbf{Low}} 
        & "Okay. Let's keep going." 
        & "Thanks for sharing that." \\
        & \textit{Explanation}: Even when emotional intensity is low, the counselor maintains engagement, which contributes to dyadic co-regulation. 
        & \textit{Explanation}: Like a conversational endpoint, with no guidance or co-regulation of emotional flow. \\
    \bottomrule
    \end{tabular}
    }
    \vspace{0.3em}
    \caption{Illustrative Counselor Utterances Comparison at Varying Arousal Levels}
    \vspace{0.5em}
    \label{tab:counselor_comparison_arousal}
\end{table*}
\subsubsection{Emotion Rise Rate}
We observed that synthetic speakers tend to respond more quickly and intensely to emotional situations than real speakers in dominance (synthetic 0.033 vs. real 0.031) and valence (synthetic 0.026 vs. real 0.024). 
In particular, dominance showed relative strong significant differences in 70\% of comparisons with the median p-value $<$ 0.05 and $r$ = 0.2). 
Similarly, valence demonstrate partial significant differences (50\% of comparisons, median p-value $<$ 0.05 and $r$ = 0.19). 
However, real speakers seem to be slightly higher arousal scores than synthetic ones with no significant differences. 

Real and synthetic counselors have very similar scores in valence (0.033 vs. 0.031), arousal (0.025, 0.026), and dominance (0.0247, 0.0254), indicating that they react to emotional situations with similar intensity and speed. 
There were no consistence significant differences observed. 

Synthetic clients seem to be more actively respond to emotional situations than real ones in the dimensions of valence (0.035, 0.032) and dominance (0.026, 0.024). 
Both dimensions showed significant differences in 60\% of comparisons with the median p-value valence $<$ 0.05 and $r$ = 0.197 and dominance $<$ 0.05 and $r$ = 0.19. 
Both real and synthetic clients reacted similarly in arousal (0.027, 0.026) with no significant differences. 

\subsubsection{Emotion Recovery Rate}
In general, synthetic and real speakers showed similar ability to return to their baseline emotional states after a shift in valence (0.033 vs. 0.031) and arousal (0.025 vs. 0.026), with no significant differences observed. 
However, dominance showed partial differences: 60\% of comparisons were significant, with a median p-value $<$ 0.05 and $r$ = 0.22, suggesting that synthetic speakers may exert slightly more emotional control in this dimension.

For counselors, emotion regulation ability was comparable across all three affective dimensions: valence (real: 0.0324 vs. synthetic: 0.0304), arousal (0.0261 vs. 0.0260), and dominance (0.0250 vs. 0.0248). No significant differences were observed.

In contrast, synthetic clients exhibited stronger emotion regulation than real clients in both valence and dominance. Valence showed significant differences in all comparisons (median p-value $<$ 0.0001 and $r$ = 0.4), while dominance showed partial significance in 50\% of comparisons (median p-value $<$ 0.05 and $r$ = 0.17). No significant differences were found in arousal, suggesting similar regulation patterns in that dimension.

\subsubsection{Emotional Arc Case Study}%: Arousal in Counselor Utterances}
To complement the quantitative results, we conduct a qualitative case study to illustrate how emotional dynamics manifest in real and synthetic CBT dialogues. Our analysis spans all three dimensions (valence, arousal, and dominance) for both counselors and clients, with detailed examples included in Appendix~\ref{sec:case_study} Table~\ref{tab:counselor_comparison_valence}--\ref{tab:client_comparison_dominance}. Here, we highlight counselor arousal (Emotion Mean) as a representative case: Table~\ref{tab:counselor_comparison_arousal} presents excerpts showing how different levels of arousal appear in practice.

\changed{We selected three real and three synthetic sessions representing the highest, median, and lowest mean arousal scores among counselor utterances. 
Table~\ref{tab:counselor_comparison_arousal} provides representative examples from each, along with interpretive commentary informed by key psychotherapy theories (e.g., emotion regulation~\cite{gross1998emerging}, calibrated empathy~\cite{elliott2018therapist, elliott2013research}, affective co-regulation~\cite{butler2013emotional}, and therapeutic alliance theory~\cite{bordin1979generalizability}).}

\changed{These examples demonstrate how different levels of affective intensity appear in real versus LLM-generated therapy conversations. 
For instance, while both real and synthetic counselors express high arousal through supportive language, synthetic speakers tend to rely on generic praise (``That’s amazing progress''), which lacks co-regulatory intent and risks feeling generic or unearned. 
In contrast, real counselors often express enthusiasm through emotionally supportive reappraisals grounded in the client’s situation (e.g., ``You’ve taken a huge step today…''). This reflects emotion regulation strategies that promote client resilience. 
At medium arousal, real counselors demonstrate calibrated empathy, engaging clients with reflective questions that validate while encouraging elaboration (e.g., ``Can you walk me through what happened…?''). 
Synthetic counselors, however, tend to offer vague prompts (e.g., “Let’s explore that feeling”), which may miss therapeutic opportunities to deepen client reflection. 
In low-arousal moments, real counselors maintain interactional grounding by guiding the session forward, preserving the structure of therapeutic engagement. 
Synthetic counterparts often deliver flat acknowledgments like “Thanks for sharing,” which, while polite, offer little in terms of affective co-regulation or forward momentum. 
This qualitative analysis reinforces our quantitative findings: although LLMs can simulate affectively expressive language, their responses may lack the situational nuance, relational grounding, and interactive adaptability found in real therapeutic interactions. }
\subsubsection{Discussion}
Synthetic sessions are generally fluent and well-structured, yet the emotional expressions of speakers still differ in subtle but important ways from those observed in real therapy interactions. 
In the following, we discuss several important distinctions in the emotional dynamics of real and synthetic CBT dialogues based on our analysis.

\textit{First, synthetic dialogues tend to exhibit higher overall emotion means, especially in arousal and valence.}
This phenomenon suggests that synthetic speakers may adopt a more emotionally elevated or expressive tone. 
This could be a result of LLMs overemphasizing affective expression to appear engaged or empathetic. 
In contrast, real counselors and clients demonstrated more restrained emotional expression, aligning with the grounded and calibrated communication style expected in professional therapy.

\textit{Second, emotion variability serves as an indicator of emotional richness in dialogue. 
We observed consistently higher variability in real sessions, particularly in arousal and valence. }
This suggests that real speakers fluctuate more in their emotional tone throughout the conversation, reflecting more authentic, responsive, and context-sensitive affective expression. 
In contrast, synthetic dialogues tend to be more emotionally uniform, which may result from LLMs generating emotionally consistent responses without deeply modeling the evolving emotional context. 

Notably, this difference in variability remains even though the synthetic dataset is ten times larger than the real dataset (580 vs. 58 dialogues). 
Despite the increased diversity that might be expected from a larger sample, the real dataset still exhibits greater emotional variability. This suggests that current LLMs are struggling to replicate emotion dynamics in real therapy sessions. 

\textit{Third, displacement length analysis further supports this observation: real speakers used emotion-laden language more dynamically, especially in arousal and dominance dimensions. }
This suggests that real speakers vary their emotional emphasis in different parts of the session, whereas synthetic speakers may be more emotionally formulaic or flat in structure.

\textit{Fourth, in terms of emotional reactivity, synthetic speakers, particularly clients, exhibited faster and more intense shifts in emotion in dominance and valence. }
While this may appear to reflect emotional engagement, it might also signal exaggerated or less nuanced affective responses, possibly resulting from LLMs optimizing for expressive content over authentic regulation.

Relatedly, emotion recovery rate, reflecting emotion regulation, was largely comparable across real and synthetic sessions, though synthetic clients demonstrated stronger emotion regulation in valence and dominance. 
This pattern might again reflect stylized generation patterns that smooth or suppress emotional volatility, possibly at the cost of naturalistic variation.

Taken together, these findings suggest that synthetic dialogues approximate but do not fully replicate the emotional dynamics of real therapy conversations. 
Synthetic counselors show relatively close alignment with real counselors across most metrics, indicating that LLMs are reasonably good at simulating professional emotional tone. 
In contrast, synthetic clients diverge more notably, particularly in reactivity and regulation, which may affect the authenticity of the simulated therapeutic process. 

\changed{To explain the observed discrepancies in emotional arcs between real and LLM-generated therapy dialogues, we identify several possible underlying factors. 
First, current LLMs may not be trained on emotionally grounded or therapeutically contextualized data~\cite{chung2023challengeslargelanguagemodels}. 
General-purpose corpora lack the subtle affective trajectories and turn-by-turn emotion regulation seen in real counseling. 
Second, common prompting strategies often elicit exaggerated or uniform emotional expressions that fail to reflect the adaptive, client-contingent nature of real therapist responses~\cite{nudo2025generativeexaggerationllmsocial}. 
Third, LLMs typically lack mechanisms for maintaining contextual coherence over multiple turns, limiting their ability to simulate affective progression across a session~\cite{shu2025fluentunfeelingemotionalblind}. 
Finally, real therapy involves ongoing co-regulation—therapists dynamically modulate their tone, intensity, and response based on client signals~\cite{butler2013emotional}. 
This interactive nuance is missing in current LLM outputs, which treat utterances as isolated generations. }

\changed{To mitigate these limitations, future work should explore fine-tuning LLMs on real counseling data to better capture authentic emotional arcs. 
Prompting strategies can also be refined to encourage adaptive rather than exaggerated affect. 
Moreover, integrating affect-aware generation and explicit co-regulatory modeling could allow LLMs to simulate more human-like emotional responsiveness and interactional flow. 
These directions offer promising steps toward enhancing the emotional fidelity of synthetic therapy data, which is critical for trustworthy use of LLMs in mental health applications.}

\subsection{Emotional Arc Similarity across Real and Synthetic Speakers}

\changed{In this section, we investigate how closely the emotional trajectories of LLM-generated speakers  align with those of real individuals. }

\changed{\subsubsection{Correlation Analysis}
Building on the method by Vishnubhotla et al.~\cite{vishnubhotla2024emotion}, we temporally align emotion arcs and compute Spearman correlations between pairs of sessions.} 
\changed{To capture different patterns of alignment, we include three pair types for each speaker role: Real–Real, Syn–Syn, and Real–Syn. 
This allows us to compare emotional arc similarity within real sessions, within synthetic sessions, and across real and synthetic sessions. 
Higher correlation values reflect stronger similarity in emotional progression. 
We report these comparisons separately for clients and counselors to highlight role-specific alignment dynamics (Table~\ref{tab:client_counselor_alignment}).}

\par\vspace{1em}  
\noindent\makebox[\linewidth]{%
\begin{minipage}{\linewidth}
    \centering
    \large % 字体更大
    \resizebox{\linewidth}{!}{%
    \begin{tabular}{p{3cm}p{2cm}p{2cm}p{2cm}}
    
    \toprule
    \textbf{Client} & \textbf{Real-Syn} & \textbf{Real-Real} & \textbf{Syn-Syn} \\
    \midrule
    Valence Mean   & 0.014  & 0.0153  & 0.2151 \\
    Arousal Mean   & 0.020  & 0.0054  & 0.0225 \\
    Dominance Mean   & 0.002  & -0.0047 & 0.0874 \\
     \midrule
    \textbf{Counselor} & \textbf{Real-Syn} & \textbf{Real-Real} & \textbf{Syn-Syn} \\
    \midrule
    Valence Mean   & 0.044  & 0.0043  & 0.1365 \\
    Arousal Mean   & -0.011 & 0.0077  & 0.0213 \\
    Dominance Mean   & 0.058  & 0.0278  & 0.1418 \\
    \bottomrule
    \end{tabular}
    }
    \vspace{0.1em}
    \captionof{table}{Correlation Analysis for Arc Alignment}
    \vspace{0.5em}
    \label{tab:client_counselor_alignment}
\end{minipage}%
}
\changed{For clients, %(Table~\ref{tab:client_counselor_alignment}), 
emotional arc similarity is weak across all pairings. 
Syn–Syn pairs show the highest values for valence (0.2151) and dominance (0.0874), but even these are modest, and arousal alignment remains negligible across all pair types. 
Both Real–Real and Real–Syn correlations are near zero or negative, indicating little consistency either within real sessions or between real and synthetic clients. 
These findings suggest that synthetic clients exhibit only limited internal consistency in their affective dynamics and fail to approximate the more complex, contingent emotion arcs of real clients. }

\changed{For counselors, %(Table~\ref{tab:client_counselor_alignment}), 
Syn–Syn correlations again yield the highest values, for valence (0.1365) and dominance (0.1418), but these are still weak. 
Real–Real and Real–Syn alignments are even lower, especially for valence and arousal. 
Notably, dominance alignment is slightly higher in Real–Syn pairs (0.058) than in Real–Real (0.0278), though the difference is small. 
Overall, these results indicate that LLM-generated counselors produce emotionally variable output, but without structured or relationally grounded patterns that resemble real therapeutic trajectories.} 
Please see the three representative examples of emotion arc correlation between real and synthetic clients in Appendix~\ref{sec:correlation_examples}.

\subsubsection{Discussion}
\changed{Our quantitative analysis of emotional arc similarity reveals that alignment is consistently low across all session pairings, including Real–Real, Syn–Syn, and Real–Syn combinations. This outcome is particularly illuminating in several ways.}

\changed{First, the low correlation between Real–Real pairs (e.g., 0.0153 for valence among clients and 0.0043 for counselors) reflects the natural heterogeneity of real-world therapy, where emotional trajectories are shaped by unique client issues, counselor styles, and the evolving goals of each session. 
In psychotherapy, affective expression is not scripted or uniform—rather, it is personalized, adaptive, and deeply contextual. 
Thus, low Real–Real correlation values serve as a baseline, not a shortcoming, and emphasize the richness and diversity of authentic human interactions.}

\changed{By contrast, LLM-generated sessions show limited internal emotional consistency (e.g., Syn–Syn valence correlations of 0.2151 for clients and 0.1365 for counselors). 
While these scores are higher than those of Real–Real or Real–Syn pairs, they remain modest overall. 
This suggests that even when generating multiple synthetic sessions, current LLMs do not produce strongly consistent emotional arcs across speakers, pointing to a lack of coherent affective modeling.}

\changed{Most notably, Real–Syn correlations are near zero across all emotion dimensions (e.g., valence = 0.014 for clients, 0.044 for counselors), indicating that synthetic speakers do not successfully emulate the affective trajectories of real speakers. 
This underscores a fundamental challenge: although LLMs can generate grammatically fluent and affectively expressive utterances, they struggle to reproduce the dynamic, situated, and role-sensitive emotional flow found in real therapy.}

\section{Conclusion}
In this work, we conduct the first systematic comparison of emotional dynamics between real and LLM-generated CBT dialogues. 
Using the UED framework, we analyzed emotion mean and variability, reactivity, regulation, and arc similarity across counselors and clients.
Our findings show that while synthetic dialogues are fluent and structured, they diverge from real therapy in key emotional dimensions---particularly in variability and emotional arc alignment. 
These results highlight the limitations of current LLMs in replicating the nuanced emotional flow of real therapeutic interactions, especially on the client side. 
This work introduces the RealCBT dataset alongside an in-depth analysis for evaluating affective realism in synthetic therapy, paving the way for future research on more emotionally grounded and psychologically credible dialogue generation.
\section{Limitations}
\changed{Our study has several limitations. 
First, RealCBT is limited in size and exhibits subgroup imbalance, which affects generalizability. 
These constraints stem from significant privacy, ethical, and legal challenges in collecting and sharing real-world CBT data. 
Unlike domains where large-scale conversational datasets are readily available, therapy transcripts are highly sensitive, and even small-scale public datasets are rare. 
Second, our analysis focuses exclusively on CBT sessions and comparisons with a single synthetic dataset (CACTUS). 
While CBT is widely practiced and structurally well-defined, making it a suitable starting point, our findings may not extend to other therapy modalities, mental health domains, or cultural and linguistic contexts. 
Nevertheless, RealCBT remains one of the few publicly available, authentic CBT datasets. 
Its inclusion enables a rare grounded comparison with synthetic dialogues and offers valuable insights into how LLM-generated therapy diverges from real interactions. 
By making RealCBT available, we hope to catalyze broader collaborative efforts toward building larger, more diverse, and ethically responsible resources that can better capture the complexity of therapeutic dialogue. }

\changed{Looking forward, we outline plans to extend research with RealCBT in several directions:
\begin{itemize}
    \item Expanding annotations to include dialogue acts, empathy markers, and turn-level emotional shifts;
    \item Training and evaluating affect-aware LLMs that model emotional dynamics more faithfully;
    \item Analyzing therapeutic strategies (e.g., reappraisal, validation) and their relationship to client emotional trajectories. 
    \item Broadening scope across therapeutic approaches (e.g., MI, psychodynamic therapy), languages, and cultural contexts, while incorporating alternative synthetic datasets to test robustness across generative sources.
\end{itemize}}

\changed{Together, these steps aim to strengthen the dataset’s utility and support progress toward building LLM systems that are both emotionally grounded and clinically meaningful.}

% Bibliography entries for the entire Anthology, followed by custom entries
%\bibliography{anthology,custom}
% Custom bibliography entries only

%\bibliographystyle{acl_natbib} 
%\bibliography{references}

    % 仅使用已生成的 bbl，避免重复文献

\clearpage
\appendix
\section{Appendix}
\label{sec:appendix}
\subsection{RealCBT Selection Criteria}
\label{sec:realcbt_criteria}
We used keywords such as \textit{CBT counseling}, \textit{CBT case study}, \textit{CBT role play}, \textit{CBT session}, and \textit{using CBT} to guide the search. 
To ensure that the selected videos accurately portrayed CBT-style therapy conversations, we applied the following inclusion criteria: (1) the video must feature exactly two participants—a counselor and a client; (2) the video should contain little to no narration or background music that could interfere with the dialogue; and (3) the session should focus on a behavioral or emotional issue consistent with CBT practice, such as depression, anxiety, or smoking cessation; (4) the therapy dialogue should last at least three minutes to ensure sufficient conversational context for analysis.

\subsection{RealCBT Summary Statistics and Distributions}
\label{sec:realcbt_summary}
Table~\ref{tab:session_stats_76} provides an overview of the RealCBT dataset, while Table~\ref{tab:distribution_realcbt_76} presents its distribution across major categories. 
To highlight key patterns, Table~\ref{tab:session_stats_58} further summarizes the top three client issues represented in the dataset, and Table~\ref{tab:distribution_realcbt_58} shows their detailed distributions.  
Together, these tables offer a comprehensive view of both the overall dataset and the prevalence of the most common client problems.  

\subsection{UED Metrics in VAD across RealCBT and CACTUS}
\label{sec:ued_metrics}
We adapt the Utterance Emotion Dynamics (UED) framework to compute emotion metrics from sequences of utterances. 
Table~\ref{tab:vad-stats} reports the mean and variability of valence, arousal, and dominance (VAD) across RealCBT and CACTUS. 
More detailed results for each dimension are presented in Table~\ref{tab:valence-stats} through Table~\ref{tab:dominance-stats}. 
Together, these statistics highlight systematic differences between real and synthetic therapy dialogues, providing a basis for the comparative analyses in the paper. 

\subsection{Representative Examples of Emotional Arcs}
\label{sec:case_study}
This appendix provides representative examples of how emotional dynamics manifest in real and synthetic CBT dialogues. The examples span all three affective dimensions (valence, arousal, and dominance) for both counselors and clients. Selected cases, shown in Table~\ref{tab:counselor_comparison_arousal} and Appendix Tables~\ref{tab:counselor_comparison_valence}--\ref{tab:client_comparison_dominance}, illustrate how real and synthetic utterances differ in affective expression across varying intensity levels.

\subsection{Representative Real–Synthetic Client Emotion Arc Correlation Examples}
\label{sec:correlation_examples}
\begin{figure*}[t] 
\centering
  \includegraphics[width=1\textwidth]{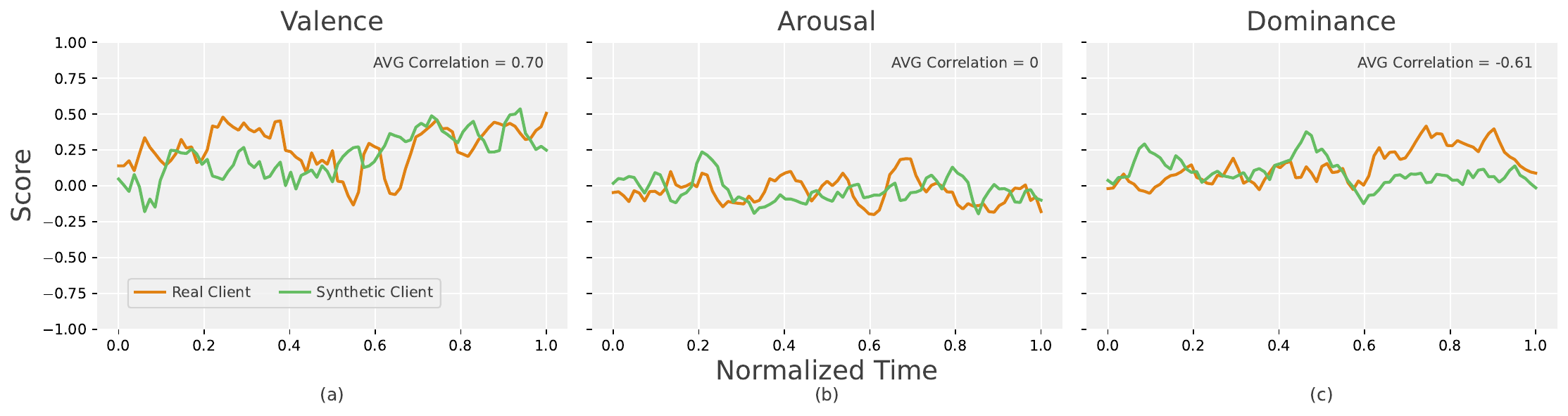}
  \caption{Emotion arcs of valence, arousal, and dominance for a client in three representative cases: (a) highest correlation, (b) near-zero correlation, and (c) lowest (negative) correlation between real and synthetic trajectories.}
  \label{fig:q2_correlation_example}
\end{figure*}

Figure~\ref{fig:q2_correlation_example} illustrates three representative examples of emotion arc correlation between real and synthetic clients:
(a) high positive correlation (strong alignment),
(b) near-zero correlation (no alignment), and
(c) high negative correlation (strong misalignment).

These examples illustrate the range of relational patterns that emerge when comparing emotional trajectories across real and LLM-generated client dialogues.

%These examples demonstrate the variability in arc similarity and emphasize that strong alignment is rare in synthetic client behavior.

\begin{table*}[t]
\centering
\resizebox{0.45\textwidth}{!}{%
\begin{tabular}{|l|r|}
\hline
Number of sessions & 76 \\
\hline
Total client words count & 82,436 \\
Total counselor words count & 108,278 \\
\hline
Total word count & 190,714 \\
Average word count per session & 2,516 \\
Total Duration (min) & 1,224.67 \\
Avg. Duration (min) & 16.11 \\
\hline
\end{tabular}
}
\caption{Descriptive statistics of \textbf{RealCBT} dataset}
\label{tab:session_stats_76}
\vspace{2em}
\end{table*}

\begin{table*}[t]
\centering
\small
\renewcommand{\arraystretch}{1.4}
\begin{tabular}{llrr}
\toprule
\textbf{Category} & \textbf{Subcategory} & \textbf{Proportion (\%)} & \textbf{Count} \\
\midrule
\multirow{9}{*}{\textbf{Client's Problem}} 
    & Anxiety and fear & 32.89 & 25 \\
    & Self-esteem and confidence issues & 25.00 & 19 \\
    & Relationships (romantic, family, friendships) & 18.42 & 14 \\
    & Career and work-related concerns & 9.21 & 7 \\
    & Health-related worries & 5.26 & 4 \\
    & Academic and educational concerns & 3.95 & 3 \\
    & Financial concerns & 2.63 & 2 \\
    & Health-related Worries & 1.32 & 1 \\
    & Other miscellaneous concerns & 1.32 & 1 \\
\midrule
\multirow{3}{*}{\textbf{Client's Attitude}} 
    & Positive & 90.79 & 69 \\
    & Neutral & 6.58 & 5 \\
    & Negative & 2.63 & 2 \\
\midrule
\multirow{2}{*}{\textbf{Client's Gender}} 
    & Female & 84.21 & 64 \\
    & Male & 15.79 & 12 \\
\midrule
\textbf{Total} & & 100.00 & 76 \\
\bottomrule
\end{tabular}
\caption{Distribution of the \textbf{RealCBT} dataset}
\label{tab:distribution_realcbt_76}
\end{table*}

\clearpage  % 强制换页
\begin{table*}[t]  % t 表示尽量在页面顶部
\centering

\begin{tabular}{|l|r|}
\hline
Number of sessions & 58 \\
\hline
Total client words count & 59,580 \\
Total counselor words count & 82,618 \\
\hline
Total word count & 142,198 \\
Average word count per session & 2,461 \\
Total Duration (min) & 907.18 \\
Avg. Duration (min) & 15.64 \\
\hline
\end{tabular}

\caption{Descriptive statistics of \textbf{RealCBT} dataset (Top 3 Client Problems)}
\label{tab:session_stats_58}
\end{table*}

\begin{table*}[t]
\centering
\small
\renewcommand{\arraystretch}{1.4}
\begin{tabular}{llrr}
\toprule
\textbf{Category} & \textbf{Subcategory} & \textbf{Proportion (\%)} & \textbf{Count} \\
\midrule
\multirow{3}{*}{\textbf{Client's Problem}} 
    & Anxiety and fear & 43.10 & 25 \\
    & Self-esteem and confidence issues & 32.76 & 19 \\
    & Relationships (romantic, family, friendships) & 24.14 & 14 \\
\midrule
\multirow{3}{*}{\textbf{Client's Attitude}} 
    & Positive & 93.10 & 54 \\
    & Negative & 3.45 & 2 \\
    & Neutral & 3.45 & 2 \\
\midrule
\multirow{2}{*}{\textbf{Client's Gender}} 
    & Female & 87.93 & 51 \\
    & Male & 12.07 & 7 \\
\midrule
\textbf{Total} & & 100.00 & 58 \\
\bottomrule
\end{tabular}
\caption{Distribution of Top 3 Client Problems in the \textbf{RealCBT} dataset}
\label{tab:distribution_realcbt_58}
\end{table*}

\begin{table*}[t]  % [t] 尽量放在页面顶部
\centering
\small
\setlength{\tabcolsep}{6pt}  % 调整列间距
\renewcommand{\arraystretch}{1.2}  % 调整行高
\begin{tabular}{|c|cc|cc|cc|}
\hline
\multicolumn{1}{|c|}{\textbf{Speaker}} & \multicolumn{2}{c|}{\textbf{Valence}} & \multicolumn{2}{c|}{\textbf{Arousal}} & \multicolumn{2}{c|}{\textbf{Dominance}} \\
\hline
 & Mean & Var. & Mean & Var. & Mean & Var. \\
\hline
Real Dialogue       & 0.2334 & 0.1448 & -0.0907 & 0.1161 & 0.0509 & 0.1131 \\
Synthetic Dialogue  & 0.2761 & 0.1480 & -0.0247 & 0.1037 & 0.1130 & 0.1154 \\
\hline
Real Counselor      & 0.2302 & 0.1466 & -0.0804 & 0.1148 & 0.0614 & 0.1120 \\
Synthetic Counselor & 0.3067 & 0.1287 & -0.0240 & 0.0998 & 0.1550 & 0.1097 \\
\hline
Real Client         & 0.2344 & 0.1312 & -0.1006 & 0.1109 & 0.0346 & 0.1052 \\
Synthetic Client    & 0.2339 & 0.1518 & -0.0257 & 0.1012 & 0.0655 & 0.1043 \\
\hline
\end{tabular}
\vspace{0.5em}
\caption{Emotion Mean and Variability in Valence, Arousal, and Dominance across RealCBT and CACTUS}
\label{tab:vad-stats}
\end{table*}

\clearpage  % 强制换页
\noindent\makebox[\textwidth]{%
\begin{minipage}{0.9\textwidth}
    \centering
    \large
    \resizebox{\textwidth}{!}{%
    \begin{tabular}{lcccccc}
    \toprule
\textbf{metric//sperker type} & \textbf{Real Dialogue} & \textbf{Synthetic Dialogue} & \textbf{Real Counselor} & \textbf{Synthetic Counselor} & \textbf{Real Client} & \textbf{Synthetic Client} \\
    \midrule
emo\_mean & 0.2334 & 0.2761 & 0.2302 & 0.3067 & 0.2344 & 0.2339 \\
    emo\_std & 0.1448 & 0.1480 & 0.1466 & 0.1287 & 0.1312 & 0.1518 \\\hline
emo\_avg\_peak\_dist & 0.0762 & 0.0779 & 0.0792 & 0.0695 & 0.0718 & 0.0818 \\
emo\_avg\_disp\_length & 4.2100 & 4.2282 & 4.2996 & 3.9746 & 3.9735 & 4.0276 \\
emo\_rise\_rate & 0.0311 & 0.0332 & 0.0328 & 0.0310 & 0.0319 & 0.0349 \\
emo\_recovery\_rate & 0.0313 & 0.0325 & 0.0324 & 0.0304 & 0.0304 & 0.0369 \\\hline
    emo\_low\_peak\_dist & 0.0896 & 0.0887 & 0.0914 & 0.0816 & 0.0838 & 0.0920 \\
emo\_low\_disp\_length & 4.3939 & 4.2972 & 4.4628 & 4.3451 & 4.3002 & 4.0820 \\
emo\_low\_rise\_rate & 0.0346 & 0.0356 & 0.0352 & 0.0321 & 0.0339 & 0.0381 \\
    emo\_low\_recovery\_rate & 0.0338 & 0.0352 & 0.0343 & 0.0328 & 0.0315 & 0.0379 \\\hline
emo\_high\_peak\_dist & 0.0666 & 0.0705 & 0.0705 & 0.0637 & 0.0664 & 0.0810 \\
emo\_high\_disp\_length & 4.2266 & 4.3962 & 4.3971 & 4.0169 & 4.0794 & 4.6133 \\
emo\_high\_rise\_rate & 0.0280 & 0.0312 & 0.0303 & 0.0304 & 0.0299 & 0.0321 \\
emo\_high\_recovery\_rate & 0.0292 & 0.0301 & 0.0305 & 0.0285 & 0.0297 & 0.0367 \\
    \bottomrule
    \end{tabular}
    }
    \vspace{0.5em}
    \captionof{table}{\centering{UED Metrics in Valence across RealCBT and CACTUS}}
     \vspace{2em}
     \label{tab:valence-stats}
\end{minipage}%
}

\noindent\makebox[\textwidth]{%
\begin{minipage}{0.9\textwidth}
    \centering
    \small
    \resizebox{\textwidth}{!}{%
    \begin{tabular}{lcccccc}
    \toprule
\textbf{metric//sperker type} & \textbf{Real Dialogue} & \textbf{Synthetic Dialogue} & \textbf{Real Counselor} & \textbf{Synthetic Counselor} & \textbf{Real Client} & \textbf{Synthetic Client} \\
    \midrule
emo\_mean & -0.0907 & -0.0247 & -0.0804 & -0.0240 & -0.1006 & -0.0257 \\
emo\_std & 0.1161 & 0.1037 & 0.1148 & 0.0998 & 0.1109 & 0.1012 \\\hline
emo\_avg\_peak\_dist & 0.0618 & 0.0557 & 0.0590 & 0.0555 & 0.0616 & 0.0564 \\
emo\_avg\_disp\_length & 4.1024 & 3.6634 & 3.8828 & 3.5889 & 4.1044 & 3.5794 \\
emo\_rise\_rate & 0.0260 & 0.0252 & 0.0254 & 0.0258 & 0.0270 & 0.0262 \\
emo\_recovery\_rate & 0.0261 & 0.0254 & 0.0261 & 0.0260 & 0.0259 & 0.0262 \\\hline
emo\_low\_peak\_dist & 0.0526 & 0.0531 & 0.0516 & 0.0536 & 0.0542 & 0.0546 \\
emo\_low\_disp\_length & 4.1279 & 3.7331 & 3.8291 & 3.7313 & 4.3172 & 3.6903 \\
emo\_low\_rise\_rate & 0.0232 & 0.0244 & 0.0236 & 0.0249 & 0.0240 & 0.0248 \\
emo\_low\_recovery\_rate & 0.0231 & 0.0243 & 0.0240 & 0.0248 & 0.0248 & 0.0253 \\\hline
emo\_high\_peak\_dist & 0.0732 & 0.0605 & 0.0723 & 0.0617 & 0.0727 & 0.0634 \\
emo\_high\_disp\_length & 4.2461 & 3.7695 & 4.4102 & 3.8029 & 4.2717 & 3.8465 \\
emo\_high\_rise\_rate & 0.0290 & 0.0263 & 0.0282 & 0.0270 & 0.0302 & 0.0278 \\
emo\_high\_recovery\_rate & 0.0293 & 0.0268 & 0.0282 & 0.0274 & 0.0275 & 0.0273 \\
    \bottomrule
    \end{tabular}
    }
    \vspace{0.5em}
    \captionof{table}{{UED Metrics in Arousal across RealCBT and CACTUS}}
    \vspace{2em}
     \label{tab:arousal-stats}
\end{minipage}%
}

\noindent\makebox[\textwidth]{%
\begin{minipage}{0.9\textwidth}
    \centering
    \small
    \resizebox{\textwidth}{!}{%
    \begin{tabular}{lcccccc}
    \toprule
\textbf{metric//sperker type} & \textbf{Real Dialogue} & \textbf{Synthetic Dialogue} & \textbf{Real Counselor} & \textbf{Synthetic Counselor} & \textbf{Real Client} & \textbf{Synthetic Client} \\
    \midrule
emo\_mean & 0.0509 & 0.1130 & 0.0614 & 0.1550 & 0.0346 & 0.0655 \\
emo\_std & 0.1131 & 0.1154 & 0.1120 & 0.1097 & 0.1052 & 0.1043 \\\hline
emo\_avg\_peak\_dist & 0.0608 & 0.0612 & 0.0595 & 0.0592 & 0.0588 & 0.0579 \\
emo\_avg\_disp\_length & 4.3631 & 4.0982 & 4.1802 & 4.0086 & 4.2093 & 3.7704 \\
emo\_rise\_rate & 0.0240 & 0.0255 & 0.0247 & 0.0254 & 0.0240 & 0.0262 \\
emo\_recovery\_rate & 0.0239 & 0.0256 & 0.0250 & 0.0248 & 0.0247 & 0.0263 \\\hline
emo\_low\_peak\_dist & 0.0619 & 0.0645 & 0.0614 & 0.0655 & 0.0610 & 0.0619 \\
emo\_low\_disp\_length & 4.3842 & 4.2835 & 4.2933 & 4.3020 & 4.3629 & 4.0320 \\
emo\_low\_rise\_rate & 0.0247 & 0.0260 & 0.0243 & 0.0268 & 0.0237 & 0.0269 \\
emo\_low\_recovery\_rate & 0.0240 & 0.0260 & 0.0247 & 0.0252 & 0.0244 & 0.0262 \\\hline
emo\_high\_peak\_dist & 0.0628 & 0.0609 & 0.0609 & 0.0587 & 0.0605 & 0.0604 \\
emo\_high\_disp\_length & 4.5743 & 4.1729 & 4.4043 & 4.2198 & 4.4529 & 4.0397 \\
emo\_high\_rise\_rate & 0.0237 & 0.0250 & 0.0250 & 0.0242 & 0.0242 & 0.0261 \\
emo\_high\_recovery\_rate & 0.0241 & 0.0254 & 0.0255 & 0.0247 & 0.0251 & 0.0266 \\
    \bottomrule
    \end{tabular}
    }
    \vspace{0.5em}
    \captionof{table}{{UED Metrics in Dominance across RealCBT and CACTUS}}
     \vspace{2em}
     \label{tab:dominance-stats}
\end{minipage}%
}

\clearpage  % 强制换页
\begin{table*}[t]
    \centering
    \footnotesize
    \resizebox{\textwidth}{!}{%
    \begin{tabular}{p{1.1cm}p{6.6cm}p{6.6cm}}
    \toprule
    \textbf{Valence Level} & \textbf{Real Counselor Snippet} & \textbf{Synthetic Counselor Snippet} \\
    \midrule
    \multirow{5}{*}{\textbf{High}} 
        & "Okay. And if you continue to feel like you have to be perfect in a presentation, you know, how do you think that will affect your future in this career?" 
        & "That’s great to hear, Andrew. As a starting point, let’s work on identifying and challenging these thoughts when they arise." \\
        & \textit{Explanation}: Connects the client’s concern to long-term goals, blending positive framing with constructive challenge. This reflects emotion regulation and supports the therapeutic alliance through future-oriented collaboration. 
        & \textit{Explanation}: Provides encouragement but in a generic way, lacking calibration to the client’s immediate narrative. This reduces opportunities for co-regulation.\\
    \midrule
    \multirow{6}{*}{\textbf{Medium}} 
        & "Let me ask you first. Logically, if you made a mistake in a presentation, would that make you feel like you were always going to make mistakes in presentations to follow?" 
        & "Understandably, strong feelings can be difficult to manage even when we recognize them as exaggerated." \\
        & \textit{Explanation}: Uses logical probing to reduce overgeneralization, showing calibrated empathy by validating but gently challenging the client. Builds the therapeutic alliance through guided reasoning. 
        & \textit{Explanation}: Normalizes emotions broadly but remains detached from the client’s specific context, limiting opportunities for emotion regulation or alliance building. \\
    \midrule
    \multirow{5}{*}{\textbf{Low}} 
        & "Where do you feel like this fear comes from? Is there an instance that occurred that caused you to be afraid or." 
        & "Sometimes our mind can amplify fears beyond what’s likely to happen. What might be some reasons or evidence that contradict this fear?" \\
        & \textit{Explanation}: Invites exploration of root causes, a low-valence but engaged stance that sustains interaction.
        & \textit{Explanation}: Provides cognitive reframing with minimal emotional tone. While constructive, it lacks the relational grounding of calibrated questioning. \\
    \bottomrule
    \end{tabular}
    }
    \caption{Illustrative Counselor Utterances Comparison at Varying Valence Levels}
    \label{tab:counselor_comparison_valence}
\end{table*}

\begin{table*}[t]
    \centering
    \footnotesize
    \resizebox{\textwidth}{!}{%
    \begin{tabular}{p{1.2cm}p{6.6cm}p{6.6cm}}
    \toprule
    \textbf{Dominance Level} & \textbf{Real Counselor Snippet} & \textbf{Synthetic Counselor Snippet} \\
    \midrule
    \multirow{5}{*}{\textbf{High}} 
        & "You've been wrestling for three years, and it is a sport you love participating in. Does it make sense that you said you were going to quit just a few matches into the season?" 
        & "I'm glad to hear you are open to it. Let’s continue to explore and challenge these beliefs together." \\
        & \textit{Explanation}: Directly confronts the client’s inconsistency, showing high directive control and authority in guiding the dialogue.  
        & \textit{Explanation}: Encourages collaboration but avoids strong challenge, reflecting lower dominance and shared control.\\
    \midrule
    \multirow{5}{*}{\textbf{Medium}} 
        & "Okay, so losing is not fun. No one likes to lose. But it doesn't mean you're a bad wrestler." 
        & "That sounds really challenging. It must be hard to feel like you're constantly judged. Can you recall any recent instances where you felt this way?" \\
        & \textit{Explanation}: Reframes the issue and provides reassurance while still shaping the client’s perspective --- moderate counselor dominance.  
        & \textit{Explanation}: Invites the client to elaborate on experiences, giving more control to the client and reducing directive influence. \\
    \midrule
    \multirow{5}{*}{\textbf{Low}} 
        & "Let me stop you right there. What if you do lose your next match? What is the worst thing that could happen?" 
        & "Let’s break it down step by step: When you noticed people staring, what specifically did you observe, and what did you think led to those thoughts and feelings?" \\
        & \textit{Explanation}: Uses hypothetical exploration to reduce anxiety; less directive than a firm challenge, reflecting lower dominance. 
        & \textit{Explanation}: Breaks the concern into steps, but with minimal counselor control, shifting responsibility to the client -- lowest dominance. \\
    \bottomrule
    \end{tabular}
    }
    \caption{Illustrative Counselor Utterances Comparison at Varying Dominance Levels}
    \label{tab:counselor_comparison_dominance}
\end{table*}

\begin{table*}[t]
    \centering
    \footnotesize
    \resizebox{\textwidth}{!}{%
    \begin{tabular}{p{1.1cm}p{6.6cm}p{6.6cm}}
    \toprule
    \textbf{Valence Level} & \textbf{Real Client Snippet} & \textbf{Synthetic Client Snippet} \\
    \midrule
    \multirow{4}{*}{\textbf{High}} 
        & "Not in a loving way, I guess, but that you should do your best, but also get the best. And I agree with that. I think that's the right thing to think." 
        & "I can do that. I’ll keep a record of those moments and I’m looking forward to our next session." \\
        & \textit{Explanation}: Expresses agreement through reflective reasoning, positive but context-grounded.   
        & \textit{Explanation}: Optimistic and cooperative, but formulaic and less tied to personal context. \\
    \midrule
    \multirow{4}{*}{\textbf{Medium}} 
        & "No, because I said their opinion. It was just in that time and place that I was stupid." 
        & "Hi, I'm doing okay, I guess. Just really anxious about some things." \\
        & \textit{Explanation}: Self-critical and context-specific, showing moderate negative valence.   
        & \textit{Explanation}: Mild disclosure of anxiety, less intense and more generic in tone. \\
    \midrule
    \multirow{4}{*}{\textbf{Low}} 
        & "Not really? I've always disliked it, and I've had bad experiences that probably made my fear worse, but it's. It's always been somewhat of a fear of mine since I was a child." 
        & "I just had this fear that I would end up feeling deprived or unsatisfied, like I’d somehow starve because I couldn’t satisfy my cravings." \\
        & \textit{Explanation}: Reflects long-term fear, personally grounded in past experiences, conveying sustained low valence. 
        & \textit{Explanation}: Negative affect framed in exaggerated hypotheticals, less grounded in lived history. \\
    \bottomrule
    \end{tabular}
    }
    \caption{Illustrative Client Utterances Comparison at Varying Valence Levels}
    \label{tab:client_comparison_valence}
\end{table*}

\begin{table*}[t]
    \centering
    \footnotesize
    \resizebox{\textwidth}{!}{%
    \begin{tabular}{p{1.1cm}p{6.6cm}p{6.6cm}}
    \toprule
    \textbf{Arousal Level} & \textbf{Real Client Snippet} & \textbf{Synthetic Client Snippet} \\
    \midrule
    \multirow{4}{*}{\textbf{High}} 
        & "I really want to go out with y'all, but I just feel like I need to work on my. On my studies, so. I'm sorry." 
        & "I had invested a substantial amount of money, so the stress was pretty high right from the start." \\
        & \textit{Explanation}: Expresses conflict and relational tension, revealing socially grounded high arousal. 
        & \textit{Explanation}: Conveys stress directly but in a more factual, less relational manner. \\
    \midrule
    \multirow{6}{*}{\textbf{Medium}} 
        & "It's like peer pressure, sort of. But I can't say no. I feel like I have to say yes just. Just because they are my friends. I do care about them and I don't want to lose them." 
        & "Maybe I could see it as a learning experience rather than a total failure. It doesn’t cancel out my other achievements, right?" \\
        & \textit{Explanation}: Shows interpersonal anxiety and fear of rejection, reflecting moderate arousal. 
        & \textit{Explanation}: Rational reframing of failure, affectively calmer and less pressured than the real example. \\
    \midrule
    \multirow{4}{*}{\textbf{Low}} 
        & "So I just kind of want to build up my self esteem and my confidence and accepting myself." 
        & "Sure. I've been feeling really unsuccessful and criticized because I've been taking losses in the stock market. It's been affecting my mood and confidence a lot." \\
        & \textit{Explanation}: Calm, constructive goal-setting, consistent with low arousal. 
        & \textit{Explanation}: Expresses discouragement with subdued tone. \\
    \bottomrule
    \end{tabular}
    }
    \caption{Illustrative Client Utterances Comparison at Varying Arousal Levels}
    \label{tab:client_comparison_arousal}
\end{table*}

\clearpage  % 强制换页
\noindent\makebox[\textwidth]{%
\begin{minipage}{\textwidth}
    \centering
    \small
    \resizebox{\textwidth}{!}{%
    \begin{tabular}{p{1.2cm}p{6.6cm}p{6.6cm}}
    \toprule
    \textbf{Dominance Level} & \textbf{Real Client Snippet} & \textbf{Synthetic Client Snippet} \\
    \midrule
    \multirow{6}{*}{\textbf{High}} 
        & "My parents don't really have the money to pay for my college, so I need to get good grades and do well with wrestling in order to have a better chance at getting a scholarship." 
        & "Thanks for having me. Well, I’ve been struggling with feeling judged because of my tattoos. It’s really been eating at me, especially in social situations." \\
        & \textit{Explanation}: Shows strong agency and responsibility, reflecting high dominance. 
        & \textit{Explanation}: Centers on external judgment, revealing less personal control. \\
    \midrule
    \multirow{4}{*}{\textbf{Medium}} 
        & "I think about it all the time, which makes it hard for me to focus." 
        & "I think it would be helpful. I really don’t want to feel like this every time I mess up." \\
        & \textit{Explanation}: Rumination disrupts focus, showing partial loss of control. 
        & \textit{Explanation}: Expresses a wish to change, indicating tentative but emerging agency. \\
    \midrule
    \multirow{6}{*}{\textbf{Low}} 
        & "I got so mad about losing that I wasn't thinking straight." 
        & "Yes, just the other day I was at the grocery store, and I caught a few people staring at me. It made me so uncomfortable that I left without buying everything I needed." \\
        & \textit{Explanation}: Behavior dictated by anger, emotions override reasoning and control. 
        & \textit{Explanation}: Actions shaped by others’ perceptions, showing externally driven loss of control. \\
    \bottomrule
    \end{tabular}
    }
    \vspace{0.5em}
    \captionof{table}{Illustrative Client Utterances Comparison at Varying Dominance Levels}
    \vspace{2em}
    \label{tab:client_comparison_dominance}
\end{minipage}%
}

\end{document}